%% file: arxiv.tex
\documentclass{article} % For LaTeX2e
\usepackage{arxiv,times}
\iclrfinalcopy
\input{math_commands.tex}

\usepackage{hyperref}
\usepackage{url}
\usepackage{xcolor}

\usepackage{booktabs}
\usepackage{multirow}
\usepackage[table]{xcolor}
\usepackage{adjustbox}
\usepackage{wrapfig}
\definecolor{empoGray}{HTML}{555555}

\usepackage{algorithm}
\usepackage{algpseudocode}
\usepackage{capt-of}

\definecolor{empoOrange}{HTML}{B8652D}

\definecolor{empoRed}{HTML}{C43F3F}

\hypersetup{colorlinks,linkcolor={blue},citecolor={magenta},urlcolor={blue}} 

\title{One Act, Many Masks: Effect-Marginal Policy Optimization for Long-Horizon Agents}

\author{%
\textbf{Shaojin Ma}$^{1,\dagger}$,
\textbf{Ruitao Wang}$^1$,
\textbf{Pengyi Li}$^{1,*}$,
\textbf{Jie Yang}$^2$,
\textbf{Liangyu Chen}$^3$ \\
\textbf{Gangyi Zhao}$^1$,
\textbf{Hongyao Tang}$^1$,
\textbf{Yan Zheng}$^1$,
\textbf{Jianye Hao}$^{1,*}$
\\[0.5ex]
$^1$Tianjin University,
$^2$Fudan University,
$^3$Renmin University of China
}

\begin{document}

\maketitle

\begingroup
\renewcommand{\thefootnote}{\fnsymbol{footnote}}

\footnotetext[2]{%
Contact me at shaojin.ma01@gmail.com%
}

\footnotetext[1]{%
\raggedright
Correspondence to:
Pengyi Li (lipengyi@tju.edu.cn) and Jianye Hao (jianye.hao@tju.edu.cn)%
}

\endgroup

\begin{abstract}

Recent advances in multi-turn reinforcement learning (RL) have substantially improved the performance of LLM agents on complex, long-horizon interactive tasks.
However, despite progress in fine-grained credit assignment, policy optimization
still applies these learning signals through response-specific probability
ratios.
Different textual responses can induce the same executable effect, yet credit
differences arising from subsequent outcomes remain tied to their individual
textual realizations.
We identify this mismatch as \textbf{credit--update misalignment}.
Our analysis reveals that credit--text pairing directly shapes policy-update
directions, even when the policy context and executable effect are held fixed.
Motivated by this finding, we propose \textbf{E}ffect-\textbf{M}arginal
\textbf{P}olicy \textbf{O}ptimization (EMPO), which aligns policy optimization
with execution-level equivalence through an effect-level surrogate that couples
updates across alternative textual realizations.
EMPO preserves the original credit signals, integrates with both trajectory-
level and step-level credit estimators, and requires no additional critics or
rollouts.
%\cley{exp results to be appended}
Experiments on ALFWorld, WebShop, and Sokoban show that EMPO outperforms strong baselines in both task performance and learning efficiency.

\end{abstract}

\section{Introduction}

Large language models (LLMs) are increasingly deployed as interactive agents for
embodied decision making, computer use, tool use, and coding.
Unlike single-turn reasoning, these tasks require agents to repeatedly perceive,
reason, and act over extended interaction horizons, often under partial observability,
sparse feedback, and delayed rewards.
Reinforcement learning (RL), which directly improves policies through environment
feedback, has therefore become an important paradigm for post-training long-horizon
LLM agents.

Credit assignment remains a central challenge in long-horizon agent RL, where
delayed outcomes must supervise many intermediate decisions.
PPO~\citep{schulman2017proximalpolicyoptimizationalgorithms} typically relies on a learned value function for
advantage estimation, while group-based methods such as
GRPO~\citep{shao2024deepseekmathpushinglimitsmathematical} replace the critic with relative comparisons among
sampled trajectories.
Recent work further improves the granularity and reliability of credit through
local state comparisons, hierarchical or graph-structured attribution, and
action-level calibration
~\citep{feng2025group, he2026hierarchyofgroups, cheng2026beyond}.
Despite their different constructions, these methods share the goal of providing
more informative credit for individual decisions along a trajectory.

\begin{figure}[t]
\centering
\includegraphics[width=\linewidth]{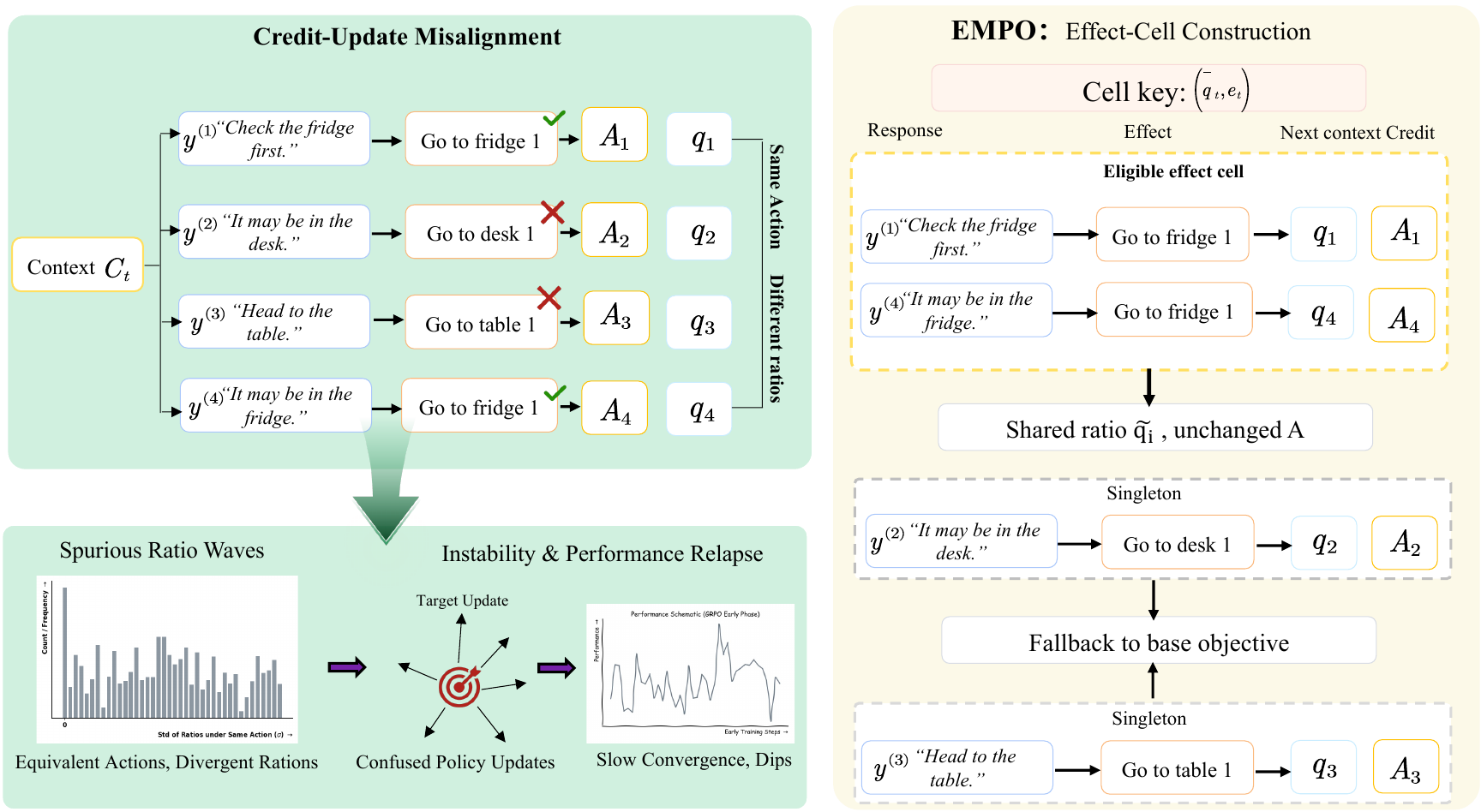}
\caption{Execution-equivalent responses may receive different credit and undergo divergent response-wise policy updates despite inducing the same immediate effect. Such within-effect variation can propagate into inconsistent ratio changes and unstable policy updates over training.}
\label{fig:credit_update_phenomenon}
\end{figure}

Yet credit estimation is only one part of policy learning, as the resulting signal
must still be translated into a policy update.
An LLM agent generates a textual response containing reasoning and an eventual
action, while an execution interface maps that response to the command that
actually affects the environment.
The mapping from textual responses to executable effects is generally many-to-one.
Responses that differ in reasoning, wording, or length may still induce the same
immediate executable effect.
Standard PPO-style objectives nevertheless retain realization-specific probability
ratios, so credit differences among execution-equivalent responses are translated
into distinct response-level probability updates.
As illustrated in Figure~\ref{fig:credit_update_phenomenon}, textual variation can
therefore introduce substantial update variation even when the resulting
environment-level decision is unchanged.

We refer to this mismatch between execution-level equivalence and
realization-specific credit--gradient pairing as
\textbf{credit--update misalignment}.
Credit assignment determines the learning signal associated with a sampled decision,
whereas policy optimization determines how that signal is coupled to changes in
the model distribution.
We find that conventional response-wise optimization contains a within-effect covariance term coupling supplied credit with realization-specific response scores.
The covariance term captures the sample-wise credit--gradient pairing retained by
independent response-level optimization, revealing an update-level source of
variation that is distinct from the quality of the underlying credit estimator.

Based on this view, we propose
\textbf{E}ffect-\textbf{M}arginal
\textbf{P}olicy \textbf{O}ptimization \textbf{(EMPO)}, which reorganizes policy
updates around executable effects rather than individual textual realizations.
EMPO identifies locally comparable responses that induce the same canonical effect
and shares their length-normalized policy ratios during optimization while retaining
the original advantage assigned to each response.
This factorizes effect-conditioned policy change from realization-specific
credit pairing without modifying the underlying credit estimator.
EMPO operates entirely on existing rollout data and introduces no additional critic,
auxiliary model, reward model, or environment interaction, making it applicable to
both trajectory-level and fine-grained credit estimators.

We evaluate EMPO on ALFWorld, WebShop, and Sokoban\citep{ALFWorld20, yao2022webshop, SchraderSokoban2018} with both GRPO and GiGPO,
covering trajectory-level and step-level credit estimation.
Across environments and model scales, EMPO improves the overall task performance
of the corresponding base algorithms.
Controlled analyses further connect the optimization objective to its intended
mechanism.
Effect-ratio sharing changes within-cell gradient coupling and leads to measurable
differences in response-probability updates after a controlled policy step, while
introducing only small additional actor-side computation.

The main contributions of this work are summarized as follows.
% We identify \textbf{credit--update misalignment}, a policy-optimization problem arising when execution-equivalent textual responses retain realization-specific credit--gradient pairing, and characterize it through rollout statistics and local gradient analysis.
(1) We identify \textbf{credit--update misalignment}, a policy-optimization
problem arising when execution-equivalent textual responses retain realization-specific credit--gradient pairing, and characterize it
through an exact decomposition of the response-wise update.
(2) We propose \textbf{E}ffect-\textbf{M}arginal
\textbf{P}olicy \textbf{O}ptimization \textbf{(EMPO)}, which shares
policy ratios across locally comparable, execution-equivalent responses while
preserving the original advantages and requiring no additional rollouts, critics,
or auxiliary models.
(3) We validate EMPO across three long-horizon agent environments, multiple model
scales, and both trajectory-level and step-level credit estimators, with controlled
analyses connecting the proposed objective to gradient coupling and realized policy
updates.

\section{Related Work}

\paragraph{Reinforcement Learning for LLM Agents.}
Reinforcement learning has become a central post-training paradigm for improving
the reasoning and decision-making capabilities of large language models.
PPO~\citep{schulman2017proximalpolicyoptimizationalgorithms} remains a foundation, while critic-free and group-relative methods such as
RLOO~\citep{arash2024rloo}, GRPO~\citep{shao2024deepseekmathpushinglimitsmathematical},
REINFORCE++~\citep{hu2026reinforce++}, and
DAPO~\citep{yu2025dapo} improve the scalability and stability of LLM reinforcement
learning.
More recently, GSPO~\citep{zheng2025groupsequencepolicyoptimization} moves importance weighting and clipping
from the token level to the sequence level.
Building on these advances, RL has increasingly been extended from single-turn
reasoning to agents that interact repeatedly with external environments.
RAGEN~\citep{ragen} studies multi-turn RL for reasoning agents in interactive
environments, while Search-R1~\citep{jin2025search},
WebAgent-R1~\citep{wei-etal-2025-webagent}, and ToolRL~\citep{qian2025toolrl}
apply RL to search, web interaction, and tool use, respectively.
Related efforts extend RL to software-engineering tasks
~\citep{wei2025swerl} and develop general interfaces for training heterogeneous
agent workflows~\citep{luo2025agentlightningtrainai}.
As agentic training moves toward longer interaction horizons, recent methods also
address challenges specific to multi-turn optimization, including adaptive
exploration~\citep{wang2026t, zhao2026aemadaptiveentropymodulation} and training stability
~\citep{li-etal-2026-turn}.
These studies establish RL as a practical framework for improving LLM agents through interaction tasks.

\paragraph{Credit Assignment in Long-Horizon Decision Making.}
Credit assignment is a longstanding challenge in reinforcement learning, where
delayed outcomes must be attributed to the intermediate decisions that produced
them.
Classical approaches reduce this difficulty through temporal value estimation,
such as generalized advantage estimation~\citep{schulman2018highdimensionalcontinuouscontrolusing}, reward shaping,
or explicit redistribution of delayed returns
~\citep{arjonamedina2019rudderreturndecompositiondelayed}.
For LLM agents, one line of work provides denser intermediate supervision through
process-level reward or verification
~\citep{choudhury2025processrewardmodelsllm, ma2026empowering}.
Another line extracts finer-grained credit directly from sampled interaction
trajectories without requiring an additional critic.
Turn-level advantage estimation assigns localized credit across multi-turn tool
interactions~\citep{wei2026reinforcingmultiturnreasoningllm}, while
GiGPO~\citep{feng2025group} constructs step-level comparison groups from recurring
environment states.
HGPO~\citep{he2026hierarchyofgroups} further accounts for historical-context
consistency when constructing these comparisons.
Other approaches exploit richer rollout structures.
SALT~\citep{li-etal-2026-salt} derives step-level advantages from trajectory graphs,
Tree-GRPO~\citep{ding2026treegrpo} obtains local supervision from tree-structured
rollouts, and GraphGPO~\citep{cheng2026beyond} estimates progress using a shared
state-transition graph.
Recent work further studies the reliability and granularity of these estimates.
ECPO~\citep{li2026densercreditenoughevidencecalibrated} aggregates returns over canonical actions and calibrates
low-support step-level credit, while GACA~\citep{liang2026granularityadaptivecreditassignmentlonghorizon} adaptively balances
trajectory- and step-level advantages according to policy uncertainty.
These methods primarily improve what credit is assigned, or how it is estimated,
densified, and calibrated. Our work studies a complementary problem at the
policy-update stage. EMPO preserves the underlying credit signal and instead
changes how it is translated into policy updates when multiple textual responses
share the same executable effect.

\section{Preliminaries}

\paragraph{Long-Horizon LLM Agent RL.}
We consider an LLM agent interacting with an environment over $T$ steps to complete
a task specified by an instruction $x$.
At interaction step $t$, the environment provides an observation $o_t$, which may
only partially reveal its underlying state.
The agent constructs a decision context $c_t$ from the task instruction and the
available interaction history, and generates a textual response
$y_t \sim \pi_\theta(\cdot \mid c_t)$.
An execution interface maps the generated response to an environment action $a_t$,
after which the environment transitions and returns a reward $r_t$ together with
the next observation $o_{t+1}$.
A complete episode produces an interaction trajectory
$\tau=\{(c_t,y_t,a_t,r_t)\}_{t=1}^{T}$.
Depending on the observability of the environment, this interaction can be viewed
as an MDP or POMDP; our formulation does not require the agent to access the
underlying environment state.

\paragraph{Group-Based Policy Optimization.}
For each task, group-based RL methods sample multiple interaction trajectories and estimate relative advantages from the resulting rollout group.
Let $\hat{A}_{i,t}$ denote the advantage assigned to step $t$ of trajectory $\tau_i$ by an arbitrary advantage estimator.
The estimated advantages are then used to optimize the LLM policy.
Our work focuses on how these credit signals are translated into policy updates, while leaving the rollout procedure, reward function, and underlying advantage estimator unchanged.

\begin{figure*}[t]
\centering
\includegraphics[width=\textwidth]{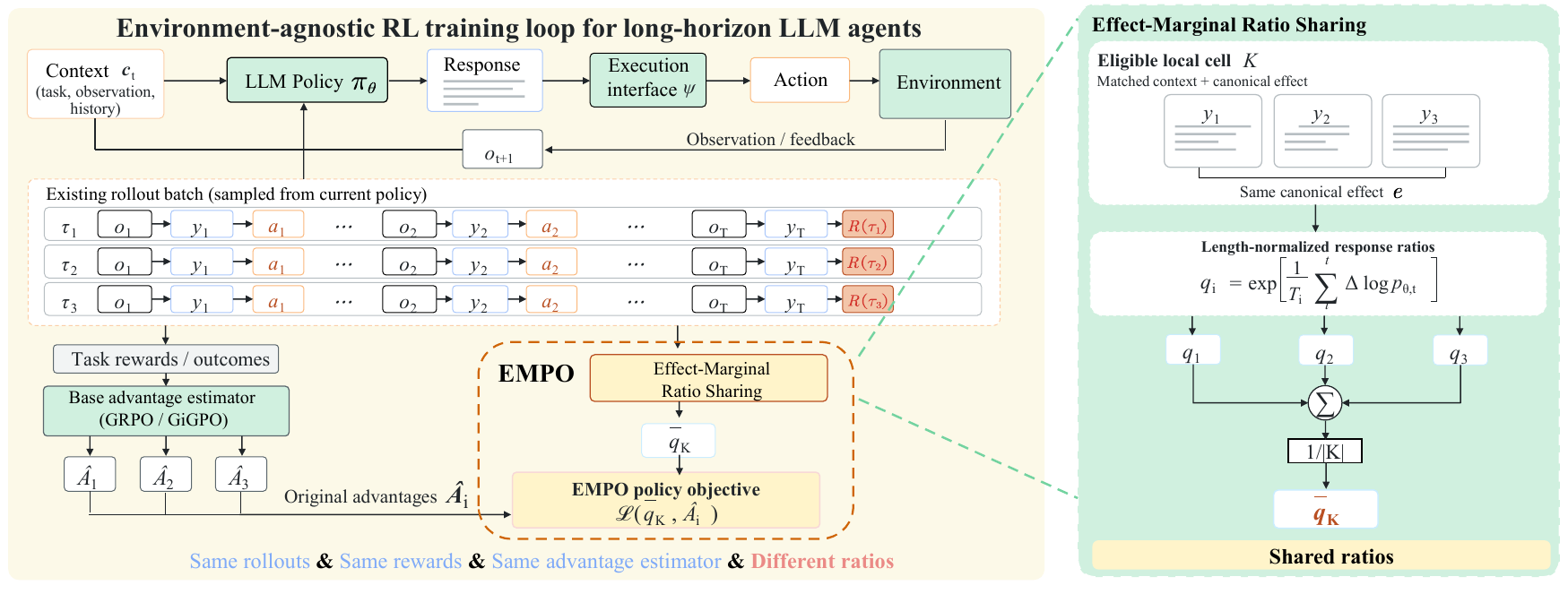}
\caption{
\textbf{Overview of Effect-Marginal Policy Optimization (EMPO).}
EMPO groups locally comparable responses with the same canonical execution effect
and aggregates their response-level policy ratios into a shared effect ratio.
The shared ratio is combined with the original advantages to coordinate policy
updates across execution-equivalent responses.
}
\label{fig:empo_overview}
\end{figure*}

% ============================================================
\section{Effect-Marginal Policy Optimization}
\label{sec:empo}

We introduce Effect-Marginal Policy Optimization (EMPO), illustrated in
Figure~\ref{fig:empo_overview}.
EMPO operates at the policy-update stage, organizing probability changes across
execution-equivalent responses while preserving the credit estimates supplied by base algorithm.
We first characterize credit--update misalignment in conventional response-wise
optimization, then derive a policy direction associated with an executable effect,
and finally instantiate this direction in a PPO-style policy objective.

% ------------------------------------------------------------
\subsection{Credit--Update Misalignment}
\label{sec:credit_update_misalignment}

At each interaction step, an LLM policy generates a textual response
$y\sim\pi_\theta(\cdot\mid c)$ under decision context $c$.
A fixed execution interface maps the response to an executed effect
\begin{equation}
    e=\psi(c,y),
\end{equation}
such as a parsed environment action or a structured tool call.
Responses with different reasoning traces, wording, or generation lengths may
therefore induce the same executable effect.

We consider a local group $K$ containing responses with comparable decision
contexts and the same executed effect.
Let $n=|K|$ and let $\hat A_i$ denote the advantage assigned to response $i$ by
the base algorithm.
As policy changes are compared and aggregated across textual realizations with
potentially different generation lengths, raw sequence-level changes introduce a
length-dependent scale.
We therefore characterize each response on a per-token scale.
For response $y_i$ of length $T_i$, define its normalized policy score at the
old-policy point as
\begin{equation}
    s_i
    =
    \left.
    \nabla_\theta
    \left[
        \frac{1}{T_i}
        \log \pi_\theta(y_i\mid c_i)
    \right]
    \right|_{\theta=\theta_{\mathrm{old}}}.
    \label{eq:normalized_policy_score}
\end{equation}
We further write
$\bar A_K=n^{-1}\sum_{i\in K}\hat A_i$
and
$\bar s_K=n^{-1}\sum_{i\in K}s_i$.

Under equal response weights and at the old-policy point, the conventional
response-wise update over $K$ is
\begin{equation}
    G_K^{\mathrm{ind}}
    =
    \frac{1}{n}
    \sum_{i\in K}
    \hat A_i s_i.
    \label{eq:ind_update}
\end{equation}
The update retains the sample-matched product between each supplied credit
$\hat A_i$ and its corresponding policy direction $s_i$.
It admits the exact decomposition
\begin{equation}
    G_K^{\mathrm{ind}}
    =
    \underbrace{
        \bar A_K\bar s_K
    }_{\text{group-level factorized component}}
    +
    \underbrace{
        \widehat{\operatorname{Cov}}_K(\hat A,s)
    }_{\text{credit--policy-direction covariance}}
    \label{eq:base_factorization}
\end{equation}
The covariance term captures how within-group variation in supplied credit
co-varies with variation in realization-specific policy directions.
Equation~\ref{eq:base_factorization} shows how this variation enters
response-wise policy updates through sample-specific credit--direction products.
We call this discrepancy between execution-level equivalence and response-wise
policy updating as \textbf{credit--update misalignment}.

% ------------------------------------------------------------
\subsection{Effect-Conditioned Policy Direction}
\label{sec:effect_direction}

We next characterize the policy direction associated with an executable effect.
For a fixed decision context $c$, the response policy induces a distribution over
executed effects:
\begin{equation}
    \pi_\theta^E(e\mid c)
    =
    \sum_{y:\psi(c,y)=e}
    \pi_\theta(y\mid c).
    \label{eq:effect_policy}
\end{equation}
Its score satisfies
\begin{equation}
    \nabla_\theta
    \log\pi_\theta^E(e\mid c)
    =
    \mathbb{E}_{Y\sim\pi_\theta(\cdot\mid c,e)}
    \left[
        \nabla_\theta
        \log\pi_\theta(Y\mid c)
    \right].
    \label{eq:effect_score}
\end{equation}
The policy direction of an executable effect is therefore determined jointly by
its alternative textual realizations.

Guided by this identity, we use the mean policy direction
$\bar s_K$ for locally comparable responses sharing the same effect.
The corresponding factorized local update is
\begin{equation}
    G_K^{\mathrm{fact}}
    =
    \frac{1}{n}
    \sum_{i\in K}
    \hat A_i\bar s_K
    =
    \bar A_K\bar s_K
    =
    \frac{1}{n^2}
    \sum_{i,j\in K}
    \hat A_i s_j.
    \label{eq:factorized_update}
\end{equation}
Compared with Eq.~\ref{eq:ind_update}, this form replaces sample-matched
credit--direction products with cross-response combinations within the same
context--effect group.

EMPO implements this factorization through the policy ratio while retaining the
advantages supplied by the base algorithm.
This confines the intervention to the policy-update stage and allows the underlying
credit estimator to remain unchanged.
The derivations of the effect-policy identity and the local factorization are
provided in Appendix~\ref{app:empo_theory}.

% ------------------------------------------------------------
\subsection{Effect-Marginal Policy Objective}
\label{sec:empo_objective}

We instantiate the effect-conditioned direction using responses available in the
rollout batch.
Let $\mathcal B$ denote the set of responses in the current optimization batch,
and let $\mathcal V\subseteq\mathcal B$ index responses with valid, non-truncated executions.
Let $h(c)$ denote the context representation used to determine local comparability.
We define an effect cell as
\begin{equation}
    K(\tilde c,e)
    =
    \left\{
        i\in\mathcal V
        \;\middle|\;
        h(c_i)=\tilde c,\;
        \psi(c_i,y_i)=e
    \right\}.
    \label{eq:effect_cell}
\end{equation}
The context key determines which decisions are locally comparable, while the
effect key identifies their shared executable decision.
Cells containing at least two responses are eligible for aggregation.

For response $i$, let
\begin{equation}
    r_{i,t}
    =
    \frac{
        \pi_\theta(y_{i,t}\mid c_i,y_{i,<t})
    }{
        \pi_{\mathrm{old}}(y_{i,t}\mid c_i,y_{i,<t})
    }
    \label{eq:token_ratio}
\end{equation}
denote the token-level likelihood ratio.
Since EMPO aggregates policy changes across multiple textual realizations, directly
aggregating full-sequence ratios would make the result sensitive to response length,
as log-ratio changes accumulate over generated tokens.
We normalize the sequence log-ratio by generation length and define the response
ratio
\begin{equation}
    q_i
    =
    \exp
    \left[
        \frac{1}{T_i}
        \sum_{t=1}^{T_i}
        \log r_{i,t}
    \right].
    \label{eq:response_ratio}
\end{equation}
For each eligible effect cell, EMPO averages these response ratios:
\begin{equation}
    \bar q_K
    =
    \frac{1}{|K|}
    \sum_{j\in K}q_j.
    \label{eq:shared_ratio}
\end{equation}

At the old-policy point,
$q_i(\theta_{\mathrm{old}})=1$ and
\begin{equation}
    \left.
    \nabla_\theta q_i
    \right|_{\theta=\theta_{\mathrm{old}}}
    =
    s_i,
    \qquad
    \left.
    \nabla_\theta\bar q_K
    \right|_{\theta=\theta_{\mathrm{old}}}
    =
    \bar s_K.
    \label{eq:shared_ratio_gradient}
\end{equation}
The shared ratio therefore realizes the effect-conditioned policy direction in
Eq.~\ref{eq:factorized_update} within the optimization objective.

For response $i$, we define the effective ratio
\begin{equation}
    \widetilde r_{i,t}
    =
    \begin{cases}
        \bar q_{K(i)},
        &
        i\in\mathcal V
        \text{ and } |K(i)|\ge 2,
        \\[4pt]
        r_{i,t},
        &
        \text{otherwise}.
    \end{cases}
    \label{eq:effective_ratio}
\end{equation}

The EMPO loss is
\begin{equation}
    \mathcal L_{\mathrm{EMPO}}(\theta)
    =
    -\frac{1}{\sum_{i\in\mathcal B} T_i}
    \sum_{i\in\mathcal B}
    \sum_{t=1}^{T_i}
    \min\left(
        \widetilde r_{i,t}\hat A_i,\,
        \operatorname{clip}
        \bigl(
            \widetilde r_{i,t},
            1-\epsilon,
            1+\epsilon
        \bigr)\hat A_i
    \right).
    \label{eq:empo_objective}
\end{equation}

Eligible responses share the cell-level ratio while retaining their individual
advantages.
Responses outside eligible cells use the original token-level ratio.
EMPO requires no additional environment interaction, critic, or auxiliary model.
Additional theoretical results are provided in
Appendix~\ref{app:empo_theory}.

\section{Experiments}
\label{sec:experiments}

We first evaluate the task performance and learning efficiency of EMPO
on long-horizon agent benchmarks (Section~\ref{sec:main_results}).
We then ablate the three operations underlying the effect-marginal
update (Section~\ref{sec:ablation}).
Finally, we examine the resulting policy-update structure
(Section~\ref{sec:rq3}) and characterize the effect cells available
for ratio sharing throughout training
(Section~\ref{sec:effect_cell_statistics}).

\subsection{Experimental Setup}
\label{sec:experimental_setup}

\paragraph{Tasks.}
We evaluate EMPO on three long-horizon agent benchmarks:
ALFWorld~\citep{ALFWorld20}, WebShop~\citep{yao2022webshop}, and
Sokoban~\citep{SchraderSokoban2018}.
ALFWorld evaluates multi-step decision making in household environments and
contains 3,827 task instances across six categories:
Pick \& Place, Examine in Light, Clean \& Place, Heat \& Place,
Cool \& Place, and Pick Two \& Place.
WebShop requires agents to search, navigate, and purchase products in a
simulated shopping website containing over 1.1 million products and 12,000
user instructions.
Sokoban evaluates sequential visual planning through box-pushing puzzles.
We report success rate (SR) on all three benchmarks and additionally report
the average task score on WebShop to capture partial task completion.

\paragraph{Baselines.}
We compare against three groups of baselines:
(1) closed-source LLMs, including GPT-4o~\citep{openai2024gpt4ocard} and
Gemini-2.5-Pro~\citep{comanici2025gemini25pushingfrontier};
(2) prompting-based agents, including direct prompting with Qwen2.5-Instruct,
ReAct~\citep{yao2023react}, and Reflexion~\citep{shinn2023reflexion};
and (3) reinforcement learning methods, including
PPO~\citep{schulman2017proximalpolicyoptimizationalgorithms},
RLOO~\citep{arash2024rloo},
GRPO~\citep{shao2024deepseekmathpushinglimitsmathematical}, and
GiGPO~\citep{feng2025group}.
We instantiate EMPO on GRPO and GiGPO, two representative RL algorithms
with trajectory-level and step-level advantage estimation, yielding EMPO-G and EMPO-GiG.

\paragraph{Training details.}
We use Qwen2.5-1.5B-Instruct and Qwen2.5-7B-Instruct
\citep{qwen2025qwen25technicalreport} for ALFWorld and WebShop, and
Qwen2.5-VL-3B-Instruct~\citep{bai2025qwen25vltechnicalreport} for Sokoban.
For each base algorithm and its EMPO counterpart, we use the same model
initialization, rollout configuration, training budget, and hyperparameter
settings.
The rollout group size is set to 8 for all group-based RL experiments. We train and evaluate each main setting with three random seeds and report mean $\pm$ standard deviation.
Full training details are provided in
Appendix~\ref{app:experimental_details}.

\begin{table*}[t]
\centering
\caption{Performance comparison on ALFWorld and WebShop benchmarks. Results are reported as mean $\pm$ std over 3 random seeds. $\Delta$ denotes the change from the corresponding baseline.}
\label{tab:main}
\small
\setlength{\tabcolsep}{3.5pt}
\renewcommand{\arraystretch}{1.02}
\setlength{\aboverulesep}{1.5pt}
\setlength{\belowrulesep}{1.5pt}
\newcommand{\result}[2]{\ensuremath{#1^{\pm #2}}}
\newcommand{\gain}[2]{\ensuremath{\mathbf{#1}^{\pm #2}}}
\newcommand{\decline}[2]{\ensuremath{\underline{#1}^{\pm #2}}}
\resizebox{0.95\textwidth}{!}{%
\begin{tabular}{@{}l*{7}{c}@{\hspace{9pt}}cc@{}}
\toprule
\textbf{Method} & \multicolumn{7}{c}{\textbf{ALFWorld}}
& \multicolumn{2}{c}{\textbf{WebShop}} \\
\cmidrule(lr){2-8}\cmidrule(l){9-10}
& Pick & Look & Clean & Heat & Cool & Pick2 & All & Score & Success \\
\midrule
\multicolumn{10}{c}{\textbf{\textit{Closed-source models}}} \\
\specialrule{0.4pt}{1pt}{1pt}
\multicolumn{10}{@{}l}{\textbf{\textit{Prompting}}} \\[-1pt]
GPT-4o & 75.3 & 60.8 & 31.2 & 56.7 & 21.6 & 49.8 & 48.0 & 31.8 & 23.7 \\
Gemini-2.5-Pro & 92.8 & 63.3 & 62.1 & 69.0 & 26.6 & 58.7 & 60.3 & 42.5 & 35.9 \\

\midrule
\multicolumn{10}{c}{\textbf{\textit{Qwen2.5-1.5B-Instruct}}} \\
\specialrule{0.4pt}{1pt}{1pt}
\multicolumn{10}{@{}l}{\textbf{\textit{Prompting}}} \\[-1pt]
Qwen2.5 & 5.9 & 5.5 & 3.3 & 9.7 & 4.2 & 0.0 & 4.1 & 23.1 & 5.2 \\
ReAct & 17.4 & 20.5 & 15.7 & 6.2 & 7.7 & 2.0 & 12.8 & 40.1 & 11.3 \\
Reflexion & 35.3 & 22.2 & 21.7 & 13.6 & 19.4 & 3.7 & 21.8 & 55.8 & 21.9 \\
\specialrule{0.3pt}{2pt}{1pt}
\multicolumn{10}{@{}l}{\textbf{\textit{RL Training}}} \\[-1pt]
PPO & \result{64.8}{3.5} & \result{40.5}{6.9} & \result{57.1}{4.9} & \result{60.6}{6.6} & \result{46.4}{4.0} & \result{47.4}{1.9} & \result{54.4}{3.1} & \result{73.8}{3.0} & \result{51.5}{2.9} \\
RLOO & \result{88.3}{3.0} & \result{52.8}{8.6} & \result{71.0}{5.9} & \result{62.8}{8.7} & \result{66.4}{5.5} & \result{56.9}{4.7} & \result{69.7}{2.5} & \result{73.9}{5.6} & \result{52.1}{6.7} \\
\addlinespace[1.5pt]
GRPO & \result{75.0}{14.4} & \result{56.9}{21.4} & \result{71.7}{15.4} & \result{84.1}{9.7} & \result{79.5}{12.0} & \result{46.1}{19.2} & \result{69.5}{11.1} & \result{76.1}{2.0} & \result{62.8}{1.6} \\
\rowcolor{gray!10}
\textbf{EMPO-G} & \gain{93.1}{4.5} & \gain{66.7}{19.1} & \gain{81.6}{4.0} & \gain{92.3}{7.7} & \gain{83.3}{3.6} & \gain{73.8}{9.0} & \gain{83.3}{3.5} & \gain{83.1}{1.3} & \gain{70.3}{2.1} \\
\rowcolor{gray!10}
\quad\textit{$\Delta$ vs. GRPO} & $+18.1$ & $+9.8$ & $+9.9$ & $+8.2$ & $+3.8$ & $+27.7$ & $+13.8$ & $+7.0$ & $+7.5$ \\
\addlinespace[1.5pt]
GiGPO & \result{98.0}{1.7} & \result{70.8}{7.2} & \result{93.1}{6.0} & \result{89.7}{4.4} & \result{81.3}{6.3} & \result{88.1}{9.0} & \result{90.1}{2.3} & \result{86.5}{3.3} & \result{73.8}{0.6} \\
\rowcolor{gray!10}
\textbf{EMPO-GiG} & \gain{99.0}{1.7} & \gain{75.0}{12.5} & \result{93.1}{0.0} & \gain{100.0}{0.0} & \gain{91.7}{3.6} & \gain{92.9}{3.6} & \gain{94.0}{1.2} & \gain{87.9}{1.4} & \gain{77.3}{3.3} \\
\rowcolor{gray!10}
\quad\textit{$\Delta$ vs. GiGPO} & $+1.0$ & $+4.2$ & $0.0$ & $+10.3$ & $+10.4$ & $+4.8$ & $+3.9$ & $+1.4$ & $+3.5$ \\
\midrule
\multicolumn{10}{c}{\textbf{\textit{Qwen2.5-7B-Instruct}}} \\
\specialrule{0.4pt}{1pt}{1pt}
\multicolumn{10}{@{}l}{\textbf{\textit{Prompting}}} \\[-1pt]
Qwen2.5 & 33.4 & 21.6 & 19.3 & 6.9 & 2.8 & 3.2 & 14.8 & 26.4 & 7.8 \\
ReAct & 48.5 & 35.4 & 34.3 & 13.2 & 18.2 & 17.6 & 31.2 & 46.2 & 19.5 \\
Reflexion & 62.0 & 41.6 & 44.9 & 30.9 & 36.3 & 23.8 & 42.7 & 58.1 & 28.8 \\
\specialrule{0.3pt}{2pt}{1pt}
\multicolumn{10}{@{}l}{\textbf{\textit{RL Training}}} \\[-1pt]
PPO & \result{92.3}{4.0} & \result{64.0}{8.4} & \result{92.5}{2.4} & \result{89.5}{7.0} & \result{80.3}{2.0} & \result{68.8}{8.3} & \result{80.4}{2.7} & \result{81.4}{3.1} & \result{68.7}{5.1} \\
RLOO & \result{87.6}{4.3} & \result{78.2}{8.3} & \result{87.3}{5.8} & \result{81.3}{7.6} & \result{71.9}{5.2} & \result{48.9}{8.4} & \result{75.5}{4.6} & \result{80.3}{3.2} & \result{65.7}{4.0} \\
\addlinespace[1.5pt]
GRPO & \result{94.0}{4.5} & \result{53.0}{29.9} & \result{88.2}{4.8} & \result{57.8}{22.7} & \result{62.6}{22.2} & \result{56.5}{7.5} & \result{71.9}{9.0} & \result{84.4}{1.8} & \result{75.8}{3.4} \\
\rowcolor{gray!10}
\textbf{EMPO-G} & \gain{94.8}{1.8} & \gain{62.5}{12.5} & \decline{77.1}{11.0} & \decline{50.0}{17.3} & \gain{75.9}{14.0} & \gain{85.7}{6.2} & \gain{80.2}{5.2} & \gain{88.1}{3.9} & \gain{77.3}{3.5} \\
\rowcolor{gray!10}
\quad\textit{$\Delta$ vs. GRPO} & $+0.8$ & $+9.5$ & $-11.1$ & $-7.8$ & $+13.3$ & $+29.2$ & $+8.3$ & $+3.7$ & $+1.5$ \\
\addlinespace[1.5pt]
GiGPO & \result{97.1}{2.9} & \result{78.6}{6.2} & \result{100.0}{0.0} & \result{73.3}{23.1} & \result{92.6}{8.5} & \result{94.8}{1.8} & \result{93.5}{2.3} & \result{85.1}{2.6} & \result{72.1}{1.8} \\
\rowcolor{gray!10}
\textbf{EMPO-GiG} & \gain{98.0}{3.4} & \gain{95.2}{8.2} & \decline{97.1}{5.0} & \gain{96.3}{6.4} & \decline{84.6}{8.5} & \decline{93.3}{6.5} & \gain{94.5}{2.8} & \gain{89.5}{3.0} & \gain{79.7}{3.4} \\
\rowcolor{gray!10}
\quad\textit{$\Delta$ vs. GiGPO} & $+0.9$ & $+16.6$ & $-2.9$ & $+23.0$ & $-8.0$ & $-1.5$ & $+1.0$ & $+4.4$ & $+7.6$ \\
\bottomrule
\end{tabular}%
}
\end{table*}

\begin{table}[t]
\centering

\makebox[\linewidth][c]{%
\begin{minipage}[b]{0.38\linewidth}
\centering

\captionof{table}{
Success rates on Sokoban with Qwen2.5-VL-3B-Instruct.
}
\label{tab:sokoban}

\small
\renewcommand{\arraystretch}{1.03}

\begin{tabular*}{\linewidth}
{@{\extracolsep{\fill}}llr@{}}
\toprule
\textbf{Type}
& \textbf{Method}
& \textbf{SR} \\
\midrule

\textit{Prompting}
& Qwen2.5-VL
& 11.7 \\

\midrule

\multirow[t]{6}{*}{\textit{RL Training}}
& GRPO
& 72.7 \\

&
\cellcolor{gray!15}\textbf{EMPO-G}
& \cellcolor{gray!15}\textbf{84.4} \\

&
\cellcolor{gray!15}\quad $\Delta$
& \cellcolor{gray!15}$+11.7$ \\

&
GiGPO
& 75.0 \\

&
\cellcolor{gray!15}\textbf{EMPO-GiG}
& \cellcolor{gray!15}\textbf{81.3} \\

&
\cellcolor{gray!15}\quad $\Delta$
& \cellcolor{gray!15}$+6.3$ \\

\bottomrule
\end{tabular*}

\end{minipage}

\hspace{0.02\linewidth}%
% ============================================================
% TTT60
% ============================================================
\begin{minipage}[b]{0.60\linewidth}
\centering
\small
\captionof{table}{
Training steps to reach 60\% success
($\mathrm{TTT}_{60}$), reported as mean $\pm$ std over three seeds.
}
\label{tab:learning_efficiency}

\small
\renewcommand{\arraystretch}{1.02}
\setlength{\tabcolsep}{2.2pt}

\begin{adjustbox}{max width=\linewidth}
\begin{tabular}{@{}cllrrr@{}}
\toprule
\textbf{Model}
& \textbf{Method}
& \textbf{Env.}
& \textbf{Base $\downarrow$}
& \textbf{+EMPO $\downarrow$}
& \textbf{Red. $\uparrow$} \\
\midrule

\multirow{4}{*}{1.5B}
& GRPO
& \multirow{2}{*}{ALFWorld}
& $108.3{\pm}15.3$
& $\mathbf{73.3{\pm}12.6}$
& $32.3\%$ \\

& GiGPO
&
& $73.3{\pm}20.2$
& $\mathbf{51.7{\pm}7.6}$
& $29.5\%$ \\

\cmidrule(l){2-6}

& GRPO
& \multirow{2}{*}{WebShop}
& $121.7{\pm}12.6$
& $\mathbf{93.3{\pm}20.8}$
& $23.3\%$ \\

& GiGPO
&
& $80.0{\pm}4.6$
& $\mathbf{73.3{\pm}12.6}$
& $8.4\%$ \\

\midrule

\multirow{4}{*}{7B}
& GRPO
& \multirow{2}{*}{ALFWorld}
& $58.3{\pm}2.9$
& $\mathbf{41.7{\pm}5.8}$
& $28.5\%$ \\

& GiGPO
&
& $36.7{\pm}7.6$
& $\mathbf{23.3{\pm}2.9}$
& $\mathbf{36.5\%}$ \\

\cmidrule(l){2-6}

& GRPO
& \multirow{2}{*}{WebShop}
& $58.3{\pm}15.3$
& $\mathbf{55.0{\pm}0.0}$
& $5.7\%$ \\

& GiGPO
&
& $50.0{\pm}5.1$
& $\mathbf{48.3{\pm}5.8}$
& $3.4\%$ \\

\bottomrule
\end{tabular}
\end{adjustbox}

\end{minipage}%
}

\end{table}

\subsection{Task Performance and Learning Efficiency}
\label{sec:main_results}

Table~\ref{tab:main} presents the main results on ALFWorld and WebShop.
EMPO consistently improves the aggregate performance of both GRPO and
GiGPO across model scales, covering trajectory-level and step-level
advantage estimation.
With Qwen2.5-1.5B-Instruct, EMPO-G improves the overall ALFWorld
success rate from 69.5\% to 83.3\% and the WebShop success rate from
62.8\% to 70.3\%.
EMPO-GiG further improves the corresponding GiGPO results from 90.1\%
to 94.0\% on ALFWorld and from 73.8\% to 77.3\% on WebShop.
The gains persist with Qwen2.5-7B-Instruct, where both EMPO variants achieve stronger overall performance on ALFWorld and WebShop.

We additionally evaluate EMPO on Sokoban with
Qwen2.5-VL-3B-Instruct.
As shown in Table~\ref{tab:sokoban}, EMPO-G improves the success rate
from 72.7\% to 84.4\%, while EMPO-GiG improves it from 75.0\% to
81.3\%.
Together, the results across ALFWorld, WebShop, and Sokoban demonstrate
consistent gains across text-based interaction and visual planning,
different model scales, and different advantage-estimation granularities.

Beyond final task performance, EMPO also improves learning efficiency.
Table~\ref{tab:learning_efficiency} reports the training steps required
to first reach a 60\% success rate ($\mathrm{TTT}_{60}$).
Across all eight evaluated model--algorithm--environment settings,
EMPO reduces $\mathrm{TTT}_{60}$, with the largest reduction reaching
36.5\%.
The corresponding learning curves are provided in
Appendix~\ref{app:training_dynamics}.

\subsection{Ablation Study}
\label{sec:ablation}

We ablate the three operations underlying EMPO, including length
normalization, effect-based grouping, and ratio sharing.
As shown in Table~\ref{tab:rq2_ablation}, all three components
contribute to the effectiveness of the effect-marginal update.
Removing length normalization makes response-level ratios increasingly
sensitive to generation length, allowing accumulated token-level
probability changes in longer responses to exert a larger influence on
the shared ratio.
Disabling ratio sharing preserves the normalized response ratio but
restores realization-specific policy directions, weakening coordinated
updates among execution-equivalent responses.
Replacing effect-based cells with arbitrary groups couples responses
without preserving executable-effect consistency, introducing
interference between policy updates associated with different effects.
These results show that length normalization calibrates policy
changes across responses, effect-based grouping determines an
appropriate coupling structure, and ratio sharing realizes coordinated
policy updates within each effect cell.

\begin{table}[t]
\centering
\caption{Ablation results using Qwen2.5-1.5B-Instruct.}
\label{tab:rq2_ablation}

\small
\setlength{\tabcolsep}{5pt}
\renewcommand{\arraystretch}{1.08}

\begin{tabular}{@{}lccc@{}}
\toprule
\textbf{Method}
& \textbf{ALFWorld SR}
& \textbf{WebShop Score}
& \textbf{WebShop SR} \\
\midrule

\rowcolor{gray!10}
\textbf{EMPO-G}
& \textbf{83.3}
& \textbf{83.1}
& \textbf{70.3} \\
\quad w/o Length Normalization
& 61.7 & 43.3 & 27.3 \\
\quad w/o Ratio Sharing
& 79.7 & 65.3 & 55.5 \\
\quad w/o Effect-Based Grouping
& 78.9 & 77.5 & 64.1 \\

\midrule

\rowcolor{gray!10}
\textbf{EMPO-GiG}
& \textbf{94.0}
& \textbf{87.9}
& \textbf{77.3} \\
\quad w/o Length Normalization
& 60.2 & 75.2 & 43.8 \\
\quad w/o Ratio Sharing
& 90.6 & 84.5 & 72.7 \\
\quad w/o Effect-Based Grouping
& 92.2 & 84.9 & 73.4 \\

\bottomrule
\end{tabular}
\end{table}

\begin{figure}[t]
\centering
\includegraphics[width=\linewidth]
{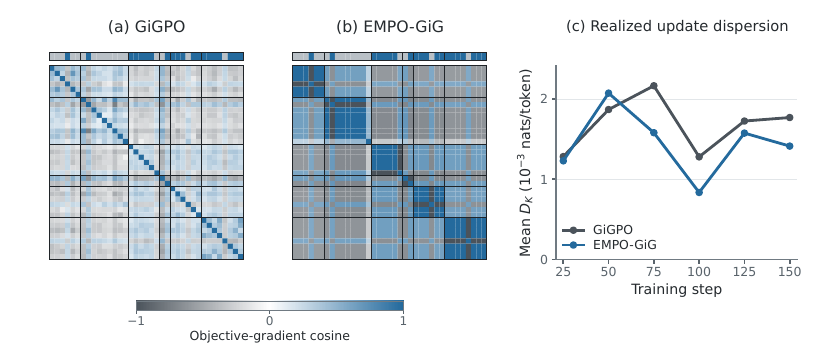}
\caption{
Policy-update analysis on ALFWorld with Qwen2.5-1.5B-Instruct.
Panels (a) and (b) show pairwise cosine similarities of response-level
objective gradients under GiGPO and EMPO-GiG. Black lines mark effect-cell
boundaries and the top strips indicate advantage signs.
Panel (c) reports the within-cell dispersion $D_K$ of log-probability
changes after controlled one-step updates.
}
\label{fig:objective_gradient_coupling}
\end{figure}

% \begin{figure}[t]
%     \centering
%     \includegraphics[width=\linewidth]
%     {figures/02_real_replies_gigpo_symlog.pdf}
%     \caption{Measured response updates under GiGPO and EMPO-GiG
%     for two execution-equivalent responses on ALFWorld.
%     Token colors show log-probability changes after a controlled
%     SGD step on a shared symmetric-log scale.
%     Reported totals are sequence log-probability changes,
%     including formatting tokens.}
%     \label{fig:response_updates_gigpo}
% \end{figure}

\subsection{Policy-Update Analysis}
\label{sec:rq3}

We further analyze how EMPO changes policy updates among
execution-equivalent responses.
Panels (a) and (b) of Figure~\ref{fig:objective_gradient_coupling}
show the pairwise cosine similarities between response-level objective
gradients.
After ratio sharing, gradients within the same effect cell exhibit a
clearer coupling structure, consistent with the factorized update in
Section~\ref{sec:empo}.

To examine the realized policy change, for each response $i$ we define
\[
\Delta_i
=
\frac{1}{T_i}
\left[
\log \pi_{\theta'}(y_i\mid c_i)
-
\log \pi_{\theta}(y_i\mid c_i)
\right],
\]
where $\theta'$ denotes the parameters after a controlled one-step
update.
For an effect cell $K$, we define the within-cell update dispersion as
$D_K=\operatorname{Std}_{i\in K}(\Delta_i)$.
Panel (c) reports the mean $D_K$ over sampled eligible cells.
EMPO-GiG yields lower within-cell dispersion overall, indicating more
coordinated realized policy updates.

\subsection{Effect-Cell Statistics}
\label{sec:effect_cell_statistics}

\begin{figure}[t]
\centering
\includegraphics[width=\linewidth]{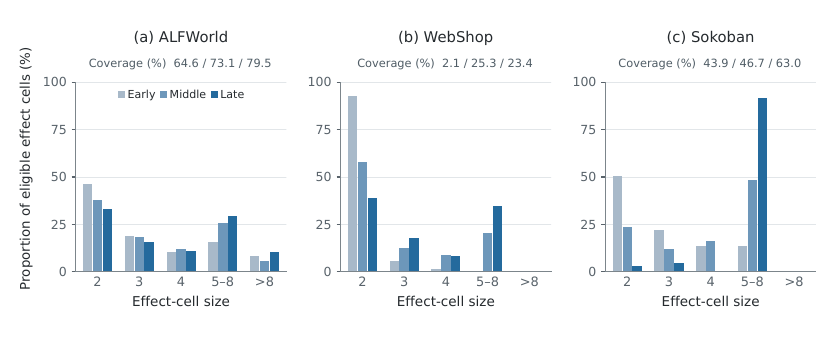}
\vspace{-12pt}
\caption{
Effect-cell size distributions of EMPO-GiG at different training stages.
Bars show the proportion of eligible cells in each size range, while
response coverage measures the fraction of sampled responses assigned to eligible cells with $|K|\geq 2$.
}
\label{fig:effect_cell_distribution}
\end{figure}

We further examine how often eligible effect cells arise during training.
Figure~\ref{fig:effect_cell_distribution} shows the effect-cell size
distribution at early, middle, and late training stages, together with the
fraction of sampled responses covered by cells with $|K|\ge2$.
Coverage increases from 64.6\% to 79.5\% on ALFWorld and from 43.9\% to
63.0\% on Sokoban, while WebShop reaches 25.3\% at the middle stage and
23.4\% at the late stage.
The corresponding size distributions also contain cells with more than two
responses, providing repeated opportunities for effect-level ratio sharing throughout training.

% \begin{wrapfigure}{r}{0.49\columnwidth}
%     \centering
%     \includegraphics[width=\linewidth]
%     {figures/training_time_donut_estimate.pdf}
%     \caption{Actor computation time (left) and peak allocated
%     GPU memory (right) for GRPO and GiGPO with and without EMPO
%     on ALFWorld and WebShop. Measurements cover actor forward,
%     loss computation, and backward passes.}
%     \label{fig:computational_overhead}
% \end{wrapfigure}

% \subsection{Computational Overhead}
% \label{sec:overhead}

% We profile the additional actor-side computation introduced by EMPO for GRPO and GiGPO on ALFWorld and WebShop. Each measurement covers forward, policy-loss computation, and backward passes over four fixed micro-batches of eight responses, with 20 paired measurements collected across two GPUs for each setting. As shown in Figure~\ref{fig:computational_overhead}, EMPO increases actor computation time by only 1.4\%--1.9\% on average across the four settings, while peak allocated GPU memory remains essentially unchanged. Since EMPO reuses existing rollouts and introduces no additional model inference, effect-level ratio sharing incurs only a small computational overhead during policy optimization.

\section{Conclusion}
\label{sec:conclusion}

We presented \textbf{E}ffect-\textbf{M}arginal
\textbf{P}olicy \textbf{O}ptimization (EMPO) for long-horizon LLM agent reinforcement learning. We identified credit--update misalignment, where execution-equivalent textual responses remain associated with realization-specific policy updates, and characterized its effect through an exact decomposition of the response-wise update.
EMPO addresses this mismatch by sharing length-normalized policy ratios
among locally comparable responses with the same executable effect while
preserving the advantages provided by the base algorithm.
Experiments on ALFWorld, WebShop, and Sokoban show that EMPO consistently
improves both task performance and learning efficiency across different
model scales and advantage estimators.
Further analysis shows that effect-level ratio sharing induces more
structured within-cell gradient coupling and more coordinated policy
updates among execution-equivalent responses.
These results demonstrate the effectiveness of effect-marginal policy
optimization for long-horizon LLM agent training.

% \section*{AI use statement}

% In this work, we did not use generative AI tools for any tasks that require disclosure. We used them only for language polishing and editing. All AI-assisted content has been reviewed, and any AI-generated text was checked and revised by the authors to ensure accuracy, originality, and technical correctness. We take full responsibility for the final content of this work, including any text, claims, or artifacts produced with the assistance of generative AI.

\bibliography{arxiv}
\bibliographystyle{arxiv}

\newpage
\appendix

\section{Algorithm}
\label{app:algorithm}

Algorithm~\ref{alg:empo} summarizes the training procedure of EMPO.

\begin{algorithm}[h]
  \caption{Effect-Marginal Policy Optimization (EMPO)}
  \label{alg:empo}
  \begin{algorithmic}[1]
  \Require Initial policy $\pi_\theta$,
  base advantage estimator $\mathcal A$

  \For{each training iteration}
      \State Freeze the behavior policy
      $\pi_{\mathrm{old}}\gets\pi_\theta$.

      \State Collect grouped trajectories with
      $\pi_{\mathrm{old}}$ and obtain environment rewards.

      \State Compute the original advantage $\widehat A_i$
      for each response using $\mathcal A$.

      \State Identify eligible effect cells by matched
      contexts $(\widetilde c_i)$ and canonical effects $(e_i)$.

      \For{each policy optimization step}
          \State Compute token ratios
          $r_{i,k}\gets
          \dfrac{\pi_\theta(y_{i,k}\mid c_i,y_{i,<k})}
          {\pi_{\mathrm{old}}(y_{i,k}\mid c_i,y_{i,<k})}$.

          \State Normalize response ratios
          $q_i\gets\exp\!\left(
              \dfrac{1}{L_i}
              \sum_{k=1}^{L_i}\log r_{i,k}
          \right)$.

          \State Initialize
          $\widetilde r_{i,k}\gets r_{i,k}$.

          \For{each eligible effect cell $K$
               participating in the update}
              \State Aggregate ratios
              $\bar q_K\gets
              \dfrac{1}{|K|}\sum_{j\in K}q_j$.

              \State Share
              $\widetilde r_{i,k}\gets\bar q_K$,
              $\quad i\in K,\ k=1,\ldots,L_i$.
          \EndFor

          \State Compute $\mathcal L_{\mathrm{EMPO}}(\theta)$
          using $\widetilde r_{i,k}$ and
          the unchanged $\widehat A_i$.

          \State Update $\theta$ with the base optimizer.
      \EndFor
  \EndFor

  \State \Return $\pi_\theta$
  \end{algorithmic}
\end{algorithm}

\section{Limitations}
\label{sec:limitations}

EMPO relies on identifying comparable interaction contexts and canonical executable effects. These are
readily available in the structured-action environments considered here, but may require more careful
construction for less structured tool interfaces. Because effect cells are formed from sampled
responses, their coverage also varies across tasks and training stages, responses outside eligible cells
retain the base objective. 
Extending effect-cell construction and evaluation to a wider range of agent interfaces is a natural
direction for future work.

\section{Theoretical Proofs}
\label{app:empo_theory}

We provide the derivations supporting the effect-conditioned policy
direction and the local factorization used by EMPO.
Throughout differentiation, sampled responses, advantages, response
lengths, and cell memberships are held fixed.

% ============================================================
\subsection{Effect-Policy Identities and Local Factorization}
\label{app:effect_factorization}

\paragraph{Effect-score identity.}
Fix a context $c$ and let
$\mathcal Y_e(c)=\{y:\psi(c,y)=e\}$.
The induced effect policy is
$\pi_\theta^E(e\mid c)
=\sum_{y\in\mathcal Y_e(c)}\pi_\theta(y\mid c)$.
Assuming that $\psi$ is independent of $\theta$, we have
\begin{align*}
\nabla_\theta\log\pi_\theta^E(e\mid c)
&=
\frac{
    \sum_{y\in\mathcal Y_e(c)}
    \nabla_\theta\pi_\theta(y\mid c)
}{
    \pi_\theta^E(e\mid c)
}
\\
&=
\sum_{y\in\mathcal Y_e(c)}
\frac{
    \pi_\theta(y\mid c)
}{
    \pi_\theta^E(e\mid c)
}
\nabla_\theta\log\pi_\theta(y\mid c)
\\
&=
\mathbb E_{Y\sim\pi_\theta(\cdot\mid c,e)}
\left[
\nabla_\theta\log\pi_\theta(Y\mid c)
\right],
\end{align*}
which proves Eq.~\ref{eq:effect_score}.

The corresponding full-response importance ratio satisfies
\[
\frac{\pi_\theta^E(e\mid c)}
     {\pi_{\mathrm{old}}^E(e\mid c)}
=
\mathbb E_{Y\sim\pi_{\mathrm{old}}(\cdot\mid c,e)}
\left[
\frac{\pi_\theta(Y\mid c)}
     {\pi_{\mathrm{old}}(Y\mid c)}
\right].
\]
EMPO uses the length-normalized response ratio
$q_i=\exp[T_i^{-1}\sum_t\log r_{i,t}]$.
We next characterize its gradient at the old-policy point.

\paragraph{Local factorization.}
Recall the normalized response score
\[
s_i
=
\left.
\nabla_\theta
\frac{1}{T_i}
\log\pi_\theta(y_i\mid c_i)
\right|_{\theta=\theta_{\mathrm{old}}},
\qquad
\bar s_K=\frac{1}{n}\sum_{i\in K}s_i.
\]
Since
$q_i(\theta_{\mathrm{old}})=1$,
\begin{equation}
\left.
\nabla_\theta q_i
\right|_{\theta_{\mathrm{old}}}
=s_i,
\qquad
\left.
\nabla_\theta\bar q_K
\right|_{\theta_{\mathrm{old}}}
=\bar s_K.
\label{eq:app_ratio_gradient}
\end{equation}

For the response-wise and shared-ratio surrogates, respectively,
\begin{align}
G_K^{\mathrm{ind}}
&=
\frac{1}{n}
\sum_{i\in K}\hat A_i s_i,
&
G_K^{\mathrm{EMPO}}
&=
\bar A_K\bar s_K
=
\frac{1}{n^2}
\sum_{i,j\in K}\hat A_i s_j.
\label{eq:app_local_gradients}
\end{align}
Using
$\bar A_K=n^{-1}\sum_i\hat A_i$,
\begin{align*}
\widehat{\operatorname{Cov}}_K(\hat A,s)
&=
\frac{1}{n}
\sum_{i\in K}
(\hat A_i-\bar A_K)(s_i-\bar s_K)
\\
&=
\frac{1}{n}
\sum_{i\in K}\hat A_i s_i
-
\bar A_K\bar s_K.
\end{align*}
Therefore,
\begin{equation}
G_K^{\mathrm{ind}}
=
G_K^{\mathrm{EMPO}}
+
\widehat{\operatorname{Cov}}_K(\hat A,s),
\label{eq:app_cov_factorization}
\end{equation}
which gives the local factorization in
Eqs.~\ref{eq:base_factorization}--\ref{eq:factorized_update}.

The same first-order relation holds for the conventional token-level
surrogate because
$\nabla_\theta r_{i,t}|_{\theta_{\mathrm{old}}}
=
\nabla_\theta\log
\pi_\theta(y_{i,t}\mid c_i,y_{i,<t})|_{\theta_{\mathrm{old}}}$.
Moreover, $r_{i,t}=1$ at this point, so PPO clipping leaves the local
gradient unchanged.

For the token-weighted loss used in practice, let
$Z=\sum_iT_i$. The corresponding cell contributions are
\[
G_K^{\mathrm{token\text{-}ind}}
=
\frac{1}{Z}
\sum_{i\in K}T_i\hat A_i s_i,
\qquad
G_K^{\mathrm{token\text{-}EMPO}}
=
\frac{\sum_{i\in K}T_i\hat A_i}{Z}\bar s_K.
\]
Applying the same decomposition to
$B_i=T_i\hat A_i$ gives
\begin{equation}
G_K^{\mathrm{token\text{-}EMPO}}
-
G_K^{\mathrm{token\text{-}ind}}
=
-\frac{n}{Z}
\widehat{\operatorname{Cov}}_K(T\hat A,s).
\label{eq:app_token_factorization}
\end{equation}

% ============================================================
\subsection{Conditional Variance}
\label{app:conditional_variance}

Fix a context--effect key $z$ and cell size $n$, and assume
$(\hat A_i,s_i)$ are conditionally i.i.d.\ given $z$.
Let
$a=\mathbb E[\hat A\mid z]$,
$\mu=\mathbb E[s\mid z]$,
$v=\operatorname{Var}(\hat A\mid z)$, and
$\Sigma=\operatorname{Cov}(s\mid z)$.

When $\hat A\perp s\mid z$, both estimators have conditional mean
$a\mu$. Expanding their second moments gives
\begin{align*}
\operatorname{Cov}
(G_K^{\mathrm{ind}}\mid z)
&=
\frac{1}{n}
\left(
a^2\Sigma
+
v\mu\mu^\top
+
v\Sigma
\right),
\\
\operatorname{Cov}
(G_K^{\mathrm{EMPO}}\mid z)
&=
\frac{1}{n}
\left(
a^2\Sigma
+
v\mu\mu^\top
\right)
+
\frac{v}{n^2}\Sigma.
\end{align*}
Hence,
\begin{equation}
\operatorname{Cov}
(G_K^{\mathrm{ind}}\mid z)
-
\operatorname{Cov}
(G_K^{\mathrm{EMPO}}\mid z)
=
\frac{n-1}{n^2}v\Sigma
\succeq0.
\label{eq:app_variance_reduction}
\end{equation}

More generally, let
$C_z=\operatorname{Cov}(\hat A,s\mid z)$.
Conditional independence across different responses gives
\[
\mathbb E[G_K^{\mathrm{ind}}\mid z]
=
a\mu+C_z,
\qquad
\mathbb E[G_K^{\mathrm{EMPO}}\mid z]
=
a\mu+\frac{1}{n}C_z.
\]
Therefore,
\begin{equation}
\mathbb E[
G_K^{\mathrm{EMPO}}
-
G_K^{\mathrm{ind}}
\mid z]
=
-\left(1-\frac{1}{n}\right)C_z.
\label{eq:app_dependence}
\end{equation}

% ============================================================
\subsection{Relation to Advantage Aggregation}
\label{app:advantage_aggregation}

For equal response weights and without clipping,
\[
\frac{1}{n}
\sum_{i\in K}\hat A_i\bar q_K
=
\bar A_K\bar q_K
=
\frac{1}{n}
\sum_{i\in K}\bar A_Kq_i.
\]
Thus, shared-ratio and cell-mean-advantage aggregation coincide for
this unclipped response-level surrogate.

PPO clipping breaks this equivalence. Let
$\phi_\epsilon(r,A)
=
\min\{rA,\operatorname{clip}(r,1-\epsilon,1+\epsilon)A\}$.
Consider
$n=2$, $\hat A_1=1$, $\hat A_2=-1$, and
$q_1=q_2=1+\delta$ with $\delta>\epsilon$.
Then $\bar A_K=0$, whereas
\[
\frac{1}{2}
\sum_{i=1}^{2}
\phi_\epsilon(\bar q_K,\hat A_i)
=
\frac{\epsilon-\delta}{2}
<0.
\]
The shared-ratio surrogate therefore retains a nonzero update in this
case, while the cell-mean-advantage surrogate vanishes.

Under the practical token-weighted reduction, define
$B_i=T_i\hat A_i$ and
$\bar B_K=n^{-1}\sum_{i\in K}B_i$.
The unclipped shared-ratio contribution satisfies
\[
\frac{1}{Z}
\sum_{i\in K}T_i\hat A_i\bar q_K
=
\frac{1}{Z}
\sum_{i\in K}
T_i
\left(
\frac{\bar B_K}{T_i}
\right)q_i.
\]
For equal response lengths,
$\bar B_K/T_i=\bar A_K$.

\section{Experimental Details}
\label{app:experimental_details}

This section provides additional details on the baselines,
environments, and training configurations used in our experiments.

% ============================================================
\subsection{Baselines}
\label{app:baselines}

\paragraph{ReAct.}
ReAct \citep{yao2023react} interleaves natural-language reasoning
and environment actions, conditioning each decision on the
observations returned by previous interactions.

\paragraph{Reflexion.}
Reflexion \citep{shinn2023reflexion} augments subsequent attempts
with natural-language reflections generated from unsuccessful
trajectories.
It provides an inference-time self-improvement baseline without
parameter updates.

\paragraph{PPO.}
Proximal Policy Optimization (PPO)
\citep{schulman2017proximalpolicyoptimizationalgorithms}
optimizes a clipped policy-gradient objective and uses a learned
value function for advantage estimation.
We include PPO as an actor--critic baseline.

\paragraph{RLOO.}
REINFORCE Leave-One-Out (RLOO)
\citep{arash2024rloo} replaces the learned critic with a
leave-one-out Monte Carlo baseline.
For $K$ trajectories with returns $\{R_i\}_{i=1}^{K}$,
\[
\hat A_i^{\mathrm{RLOO}}
=
R_i
-
\frac{1}{K-1}
\sum_{j\neq i}R_j.
\]

\paragraph{GRPO.}
Group Relative Policy Optimization (GRPO)
\citep{shao2024deepseekmathpushinglimitsmathematical}
normalizes returns among trajectories sampled for the same task.
For a trajectory group $\mathcal G$,
\[
\hat A_i^{\mathrm{GRPO}}
=
\frac{
R_i-\mu_{\mathcal G}
}{
\sigma_{\mathcal G}+\varepsilon
},
\]
where $\mu_{\mathcal G}$ and $\sigma_{\mathcal G}$ are the group
mean and standard deviation.
The resulting advantages are optimized with a PPO-style clipped
objective without a learned critic.

\paragraph{GiGPO.}
GiGPO \citep{feng2025group} augments the trajectory-level
group-relative advantage with a step-level relative advantage
computed among interactions sharing comparable decision contexts:
\[
\hat A_{i,t}^{\mathrm{GiGPO}}
=
\hat A_i^{\mathrm{traj}}
+
\omega\hat A_{i,t}^{\mathrm{step}}.
\]
We follow the original implementation with $\omega=1$ and
discount factor $\gamma=0.95$.

% ============================================================
\subsection{Environments}
\label{app:environments}

\paragraph{ALFWorld.}
ALFWorld \citep{ALFWorld20} is a text-based embodied environment
derived from ALFRED.
Each episode specifies a household task that must be completed
through a sequence of textual actions.
The benchmark contains six task categories: Pick \& Place,
Examine in Light, Clean \& Place, Heat \& Place, Cool \& Place,
and Pick Two \& Place.

\paragraph{WebShop.}
WebShop \citep{yao2022webshop} evaluates web-based decision making
in a simulated shopping environment.
The agent searches and navigates product pages, selects product
options, and completes a purchase according to a natural-language
instruction.
The environment contains more than 1.1 million products and
12,000 user instructions.

\paragraph{Sokoban.}
We use procedurally generated Sokoban instances to evaluate EMPO
in a visual planning setting.
The agent observes an RGB rendering of a $6\times6$ board with
one box and outputs movement actions to push the box onto the
target location.
Boards are generated with a room-generation search depth of 30,
and each episode is limited to 15 interactions.

\subsection{Training Metrics}
\label{app:training_metrics}

\paragraph{Policy-gradient loss.}
We report the clipped surrogate loss used to optimize the actor.
Its value is determined jointly by the estimated advantages, policy
ratios, and clipping behavior, and therefore reflects the optimization
signal encountered by the policy during training.

\paragraph{KL divergence.}
We track the token-level KL divergence between the current policy and
the fixed reference policy.
This quantity measures the accumulated deviation of the learned policy
from its initialization over the course of training.

\paragraph{Policy-ratio clip fraction.}
The clip fraction measures the proportion of policy-update terms for
which the likelihood ratio falls outside the PPO clipping interval.
It characterizes how frequently proposed probability changes are
constrained by the clipped surrogate objective.
For EMPO, this statistic is computed using the effective ratios defined
in Eq.~\ref{eq:empo_objective}.

\paragraph{Training reward and success rate.}
We report the mean episodic reward and the fraction of successful
episodes in each rollout batch to track on-policy task progress during
training.
The main experimental results are evaluated separately on the
corresponding validation sets.

\paragraph{Token entropy.}
Token entropy is computed from the next-token distribution along
sampled response prefixes and averaged over valid response tokens.
It measures the diversity of the model's token-level policy throughout
training.

\paragraph{Gradient norm.}
We record the norm of the actor gradient before gradient clipping.
This statistic characterizes the scale of the proposed parameter update
and is used to monitor optimization dynamics throughout training.
%============================================================
\subsection{Training Details}
\label{app:implementation}

We use Qwen2.5-1.5B-Instruct and Qwen2.5-7B-Instruct for
ALFWorld and WebShop, and Qwen2.5-VL-3B-Instruct for Sokoban.
All models are optimized with AdamW using an actor learning rate
of $10^{-6}$ and weight decay of $0.01$.
The critic learning rate for PPO is $10^{-5}$.

We use a PPO clipping parameter of $0.2$, dual clipping of $3.0$,
a KL-loss coefficient of $0.01$, an entropy coefficient of
$0.001$, and gradient clipping at $1.0$.
Rollouts are sampled with temperature $1.0$, while evaluation
uses temperature $0.4$.
We set top-$p$ to $1.0$ and disable top-$k$ filtering.
The rollout group size is eight for all group-based methods.

Unless otherwise stated, models are trained for 150 iterations
and evaluated every five iterations using 128 episodes.
The maximum response length is 512 tokens in all environments.
Environment-specific settings are summarized in
Table~\ref{tab:env_settings}.

\begin{table}[t]
\centering
\caption{Environment-specific training configurations.}
\label{tab:env_settings}
\small
\setlength{\tabcolsep}{5pt}
\begin{tabular}{@{}lccc@{}}
\toprule
 & ALFWorld & WebShop & Sokoban \\
\midrule
Max. prompt length
    & 2,048 & 4,096 & 1,024 \\
Max. environment steps
    & 50 & 15 & 15 \\
Groups $\times$ trajectories
    & $16\times8$ & $16\times8$ & $32\times8$ \\
Actor mini-batch size
    & 256 & 64 & 64 \\
\bottomrule
\end{tabular}
\end{table}

% ============================================================

% ============================================================

\subsection{Computing Resources}
\label{app:computing_resources}

Experiments are run on eight NVIDIA H100 GPUs.
We build EMPO on \texttt{verl-agent}\footnote{\url{https://github.com/langfengQ/verl-agent}}, using PyTorch FSDP for distributed training and vLLM for rollout generation.
Gradient checkpointing and bfloat16 mixed precision are enabled.

Rollout tensor parallelism is set to two for the 1.5B language models and the 3B vision--language model, and to four for the 7B
models. Optimizer-state offloading is enabled for the 7B experiments. EMPO uses the same actor model and environment rollouts as its
corresponding base algorithm.

\section{Further Analysis}
\label{app:further_analysis}

\subsection{Learning Curves}
\label{app:learning_curves}

\begin{figure}[t]
\centering
\includegraphics[width=\linewidth]{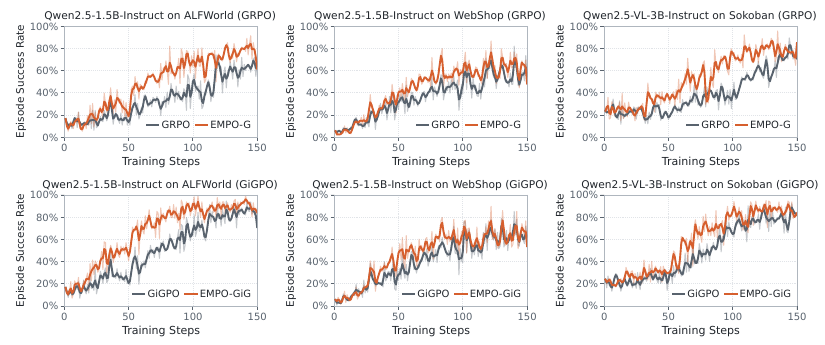}
\caption{Training curves on ALFWorld, WebShop, and Sokoban.}
\label{fig:rq1_learning_efficiency}
\end{figure}

Figure~\ref{fig:rq1_learning_efficiency} provides the learning curves
corresponding to the $\mathrm{TTT}_{60}$ results reported in
Table~\ref{tab:learning_efficiency}.
Across ALFWorld, WebShop, and Sokoban, EMPO generally reaches stronger
policies earlier in training and maintains favorable learning progress
over the training horizon.

\begin{figure}[t]
    \centering
    \includegraphics[width=\linewidth]
    {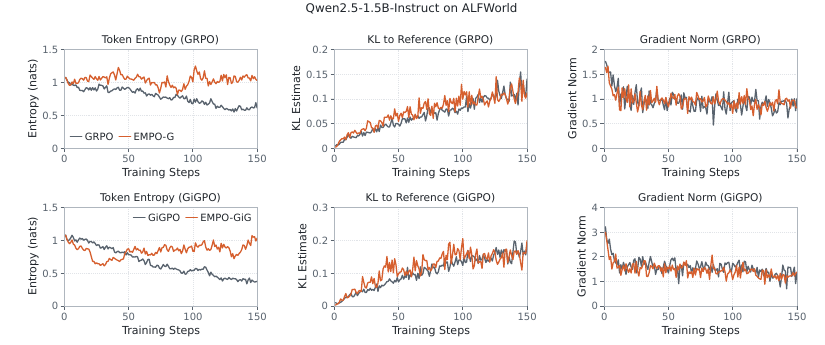}
    \caption{
    Training dynamics on ALFWorld with Qwen2.5-1.5B-Instruct.
    We report token entropy, KL divergence to the reference policy,
    and gradient norm.
    }
    \label{fig:training_dynamics_alfworld}
\end{figure}

\begin{figure}[t]
    \centering
    \includegraphics[width=\linewidth]
    {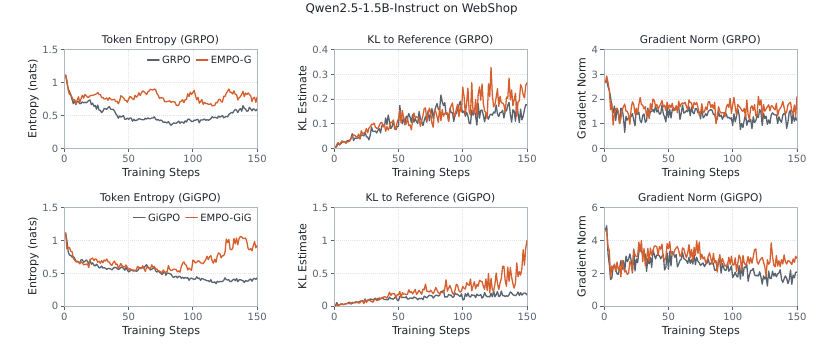}
    \caption{
    Training dynamics on WebShop with Qwen2.5-1.5B-Instruct.
    We report token entropy, KL divergence to the reference policy,
    and gradient norm.
    }
    \label{fig:training_dynamics_webshop}
\end{figure}

\subsection{Magnitude of Credit--Direction Covariance}
\label{app:covariance_magnitude}

The decomposition in Eq.~\ref{eq:base_factorization} isolates a
within-effect covariance term that captures the sample-specific pairing
between supplied credit and realization-specific policy directions.
We examine the magnitude of this term on effect cells sampled from
ALFWorld.

\begin{figure}[t]
    \centering
    \includegraphics[width=\linewidth]{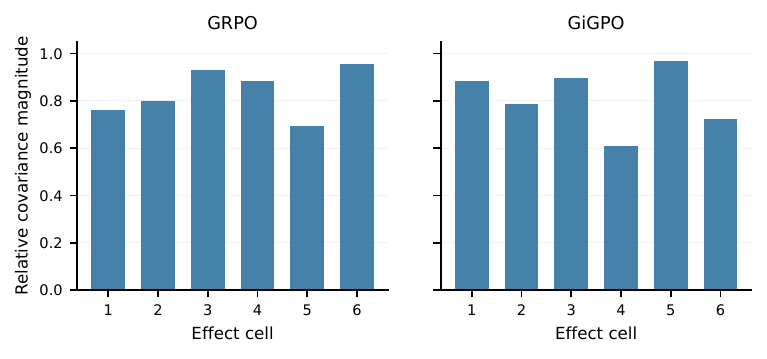}
    \caption{
    Relative magnitude of the within-effect covariance term on six
    effect cells from ALFWorld.
    For each cell $K$, we report
    $\|\widehat{\operatorname{Cov}}_K(\hat A,s)\|/
    \|G_K^{\mathrm{ind}}\|$
    under GRPO and GiGPO.
    }
    \label{fig:covariance_magnitude}
\end{figure}

For each effect cell $K$, Figure~\ref{fig:covariance_magnitude} reports
the relative covariance magnitude
$\|\widehat{\operatorname{Cov}}_K(\hat A,s)\|/
\|G_K^{\mathrm{ind}}\|$.
Across the analyzed cells, the covariance term remains comparable in
magnitude to the conventional response-wise update under both GRPO and
GiGPO.
This empirical pattern complements the gradient decomposition in
Section~\ref{sec:empo} and shows that within-effect credit--direction
covariance constitutes a substantial component of the policy update in
practice.

\subsection{Comparison with Advantage Aggregation}
\label{app:advantage_aggregation_exp}

Section~\ref{app:advantage_aggregation} shows that sharing the
response ratio and averaging the advantage coincide for an unclipped
response-level surrogate with equal response weights, while the
equivalence no longer holds under the practical clipped and
token-weighted objective.
We empirically compare EMPO with the corresponding advantage-side
aggregation variant on ALFWorld and WebShop.

\begin{table}[h]
\centering
\caption{
Comparison between EMPO and advantage-side aggregation.
}
\label{tab:advantage_aggregation_exp}

\small
\setlength{\tabcolsep}{4.5pt}
\renewcommand{\arraystretch}{1.05}

\begin{tabular}{@{}llccc@{}}
\toprule
\textbf{Base}
& \textbf{Method}
& \textbf{ALFWorld SR}
& \textbf{WebShop Score}
& \textbf{WebShop SR} \\
\midrule

GRPO
& Advantage Aggregation
& 80.47
& 82.1
& 70.2 \\

& EMPO-G
& \textbf{83.3}
& \textbf{83.1}
& \textbf{70.3} \\

\midrule

GiGPO
& Advantage Aggregation
& 92.97
& 83.6
& 71.1 \\

& EMPO-GiG
& \textbf{94.0}
& \textbf{87.9}
& \textbf{77.3} \\

\bottomrule
\end{tabular}
\end{table}

As shown in Table~\ref{tab:advantage_aggregation_exp}, the
advantage-side variant remains competitive, consistent with the
close relationship established by the unclipped analysis.
EMPO achieves higher performance in all reported settings, with the
difference becoming more pronounced for GiGPO on WebShop.
These results are consistent with the distinction derived in
Section~\ref{app:advantage_aggregation}: once clipping and practical
loss weighting are introduced, ratio-side and advantage-side
aggregation induce different policy updates.

\subsection{Policy Adaptation and Response Diversity}
\label{app:training_dynamics}

We further examine how effect-marginal optimization shapes policy
adaptation throughout training.
Figures~\ref{fig:training_dynamics_alfworld}
and~\ref{fig:training_dynamics_webshop} report token entropy,
KL divergence to the reference policy, and gradient norm for
GRPO, GiGPO, and their EMPO variants.

Across both ALFWorld and WebShop, EMPO maintains higher token entropy
during the later stages of training while exhibiting continued policy
movement, as reflected by the KL divergence and gradient norm.
The three quantities together reveal a characteristic optimization
pattern: the policy continues to adapt while retaining a less
concentrated token distribution.
This pattern appears consistently for both trajectory-level GRPO and
step-level GiGPO, indicating that it is preserved across different
advantage-estimation granularities.

This training behavior connects naturally to the effect-marginal
update in Section~\ref{sec:empo}.
In response-wise optimization, each credit signal is paired with its
realization-specific policy direction, allowing within-cell credit
variation to directly shape the probability of individual textual
realizations.
EMPO couples execution-equivalent responses through a shared
effect-level ratio, so their policy changes are organized around a
common effect-conditioned direction.
The resulting entropy dynamics provide a distribution-level view of
the coordinated updates observed in the policy-update analysis in
Section~\ref{sec:rq3}.

\section{Prompt Templates}
\label{app:prompts}

We use environment-specific prompt templates specifying the task,
available action space, interaction history, and output format.

% Prompt figures will be inserted here.

% Example:

\begin{figure*}[h]
    \centering
    \includegraphics[width=0.75\textwidth]{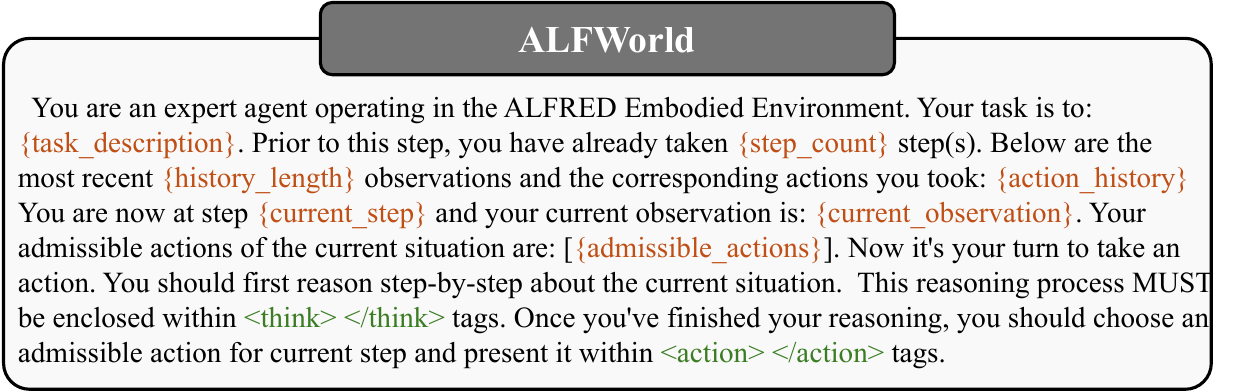}
    \caption{Prompt template used for ALFWorld.}
    \label{fig:alfworld_prompt}
\end{figure*}

\begin{figure*}[h]
    \centering
    \includegraphics[width=0.75\textwidth]{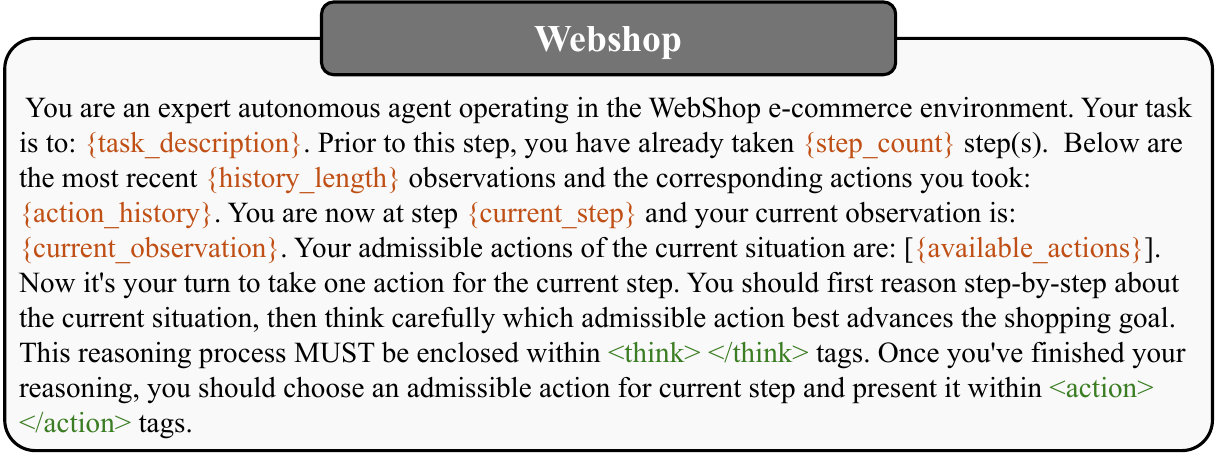}
    \caption{Prompt template used for WebShop.}
    \label{fig:webshop_prompt}
\end{figure*}

\begin{figure*}[h]
    \centering
    \includegraphics[width=0.75\textwidth]{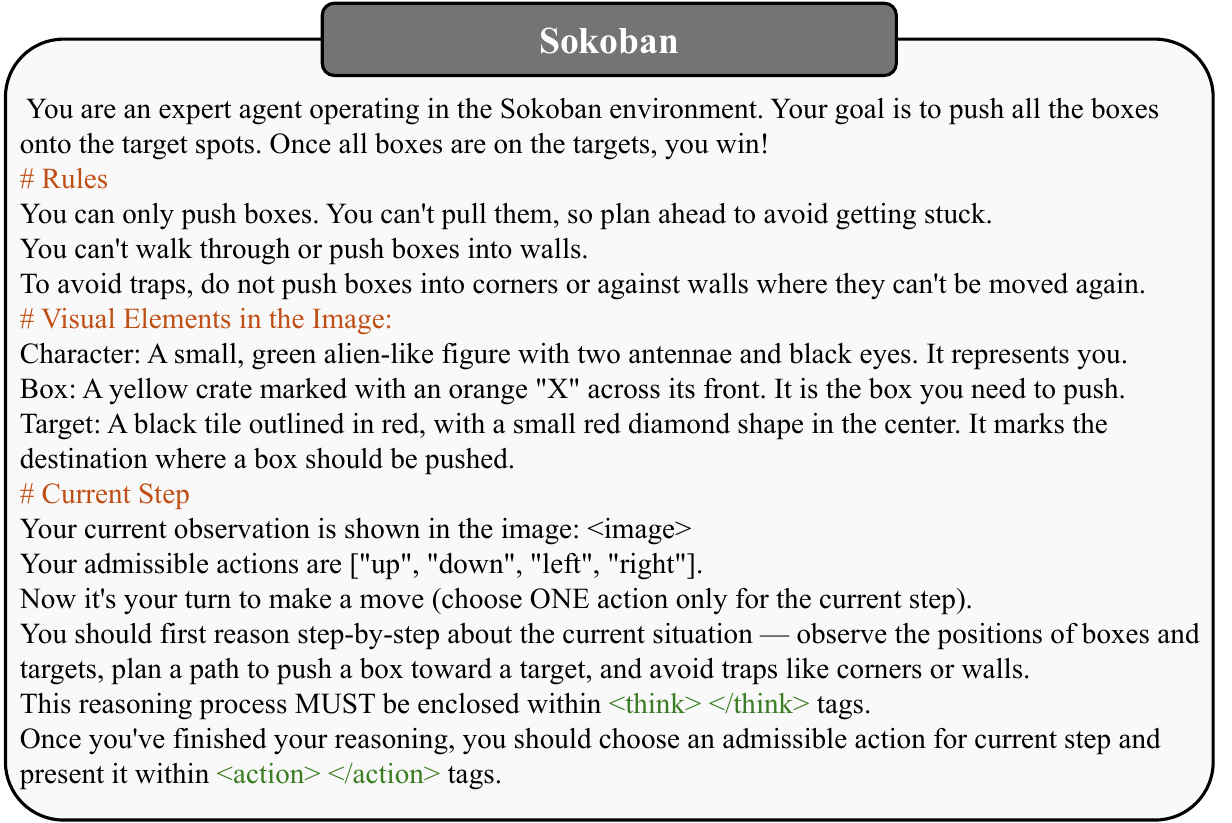}
    \caption{Prompt template used for Sokoban.}
    \label{fig:sokoban_prompt}
\end{figure*}

\end{document}

%% file: math_commands.tex
\usepackage{amsmath,amsfonts,bm}

\def\eqref#1{equation~\ref{#1}}
\def\1{\bm{1}}

\DeclareMathAlphabet{\mathsfit}{\encodingdefault}{\sfdefault}{m}{sl}
\SetMathAlphabet{\mathsfit}{bold}{\encodingdefault}{\sfdefault}{bx}{n}